\documentclass{ijcaArticle}
\ijcaVolume{VV}
\ijcaNumber{N}
\ijcaYear{YYYY}
\ijcaMonth{Month}

\usepackage[utf8]{inputenc}
\usepackage{graphicx}
\usepackage{amsmath}
\usepackage{booktabs}
\usepackage{url}
\usepackage{float}
\usepackage{cite}
\usepackage{multirow}
\usepackage{siunitx}
\usepackage{tabularx}
\usepackage{booktabs} 
\usepackage{tabularx} 
\usepackage{caption}

\begin{document}

\title{When Better Eyes Lead to Blindness: A Diagnostic Study of the Information Bottleneck in CNN-LSTM Image Captioning Models}

\author{ 
   \large Hitesh Kumar Gupta \\[-3pt]
   \normalsize Independent Researcher \\[-3pt]
   \normalsize hiteshagarwal172@gmail.com \\[-3pt]
}

\terms{Deep Learning, Computer Vision, Natural Language Processing, Multimodal Learning}

\maketitle

\begin{abstract} 
Image captioning, situated at the intersection of computer vision and natural language processing, requires a sophisticated understanding of both visual scenes and linguistic structure. While modern approaches are dominated by large-scale Transformer architectures, this paper documents a systematic, iterative development of foundational image captioning models, progressing from a simple CNN-LSTM encoder-decoder to a competitive attention-based system. This paper presents a series of five models, beginning with \textit{Genesis} and concluding with \textit{Nexus}, an advanced model featuring an EfficientNetV2B3 backbone and a dynamic attention mechanism. The experiments chart the impact of architectural enhancements and demonstrate a key finding within the classic CNN-LSTM paradigm: merely upgrading the visual backbone without a corresponding attention mechanism can degrade performance, as the single-vector bottleneck cannot transmit the richer visual detail. This insight validates the architectural shift to attention. Trained on the MS COCO 2017 dataset, the final model, \textit{Nexus}, achieves a BLEU-4 score of 31.4, surpassing several foundational benchmarks and validating the iterative design process. This work provides a clear, replicable blueprint for understanding the core architectural principles that underpin modern vision-language tasks.
\end{abstract}

\section{Introduction}
The automated generation of natural language descriptions for images, known as image captioning, remains a challenging and active area of research. It represents a classic multi-modal task, demanding that a model not only recognize objects and their attributes within an image but also understand their relationships and compose a grammatically correct and semantically rich sentence.

The dominant approach to this problem has been the encoder-decoder framework \cite{vinyals2015show}, which typically uses a Convolutional Neural Network (CNN) to encode an image into a fixed-length vector representation, and a Recurrent Neural Network (RNN) to decode this vector into a sequence of words. While pioneering, this paradigm is sensitive to a wide array of architectural choices, from the depth of the CNN backbone to the complexity of the RNN decoder and the method of feature fusion.

In this paper, we present the results of a structured, iterative research process aimed at building and refining an image captioning model. The methodology is rooted in progressive enhancement, starting with a simple baseline and methodically introducing more complex components to address specific limitations. This paper documents the development of five distinct models, named to reflect their role in this evolution: \textit{Genesis}, \textit{Contexta}, \textit{Clarity}, \textit{Focalis}, and \textit{Nexus}.

This journey provides valuable insights into the interplay between visual perception, language modeling, and attention. This paper demonstrates empirically how each architectural modification impacts performance, including a counter-intuitive but instructive finding where a more powerful visual encoder led to a decrease in caption quality, thereby motivating the shift to an attention-based design. This work also serves as a pedagogical guide for students and practitioners, offering a clear, replicable framework to understand the architectural trade-offs that shaped modern vision-language models. The final model, \textit{Nexus}, leverages the lessons learned to achieve results that are competitive with and, on several key metrics surpass, foundational benchmarks in the field.

\section{Related Work}
The field of image captioning has evolved significantly since its inception. This section reviews the foundational encoder–decoder paradigms upon which this work builds, as well as the modern Transformer-based architectures that now define the state of the art.

\subsection{Foundational Encoder-Decoder Architectures}
Early work established the now-classic encoder-decoder architecture. Vinyals et al. \cite{vinyals2015show} in their "Show and Tell" model, successfully repurposed techniques from machine translation, using a pre-trained CNN like InceptionV3 as an encoder and an LSTM network as a decoder. The initial model, \textit{Genesis}, is a direct implementation of this foundational paradigm.

The primary limitation of this approach was the information bottleneck created by compressing the entire image into a single static context vector. The breakthrough came with the introduction of visual attention mechanisms by Xu et al. \cite{xu2015show} in "Show, Attend and Tell". Their model allowed the decoder to dynamically "look" at different parts of the image at each step of the caption generation process. This concept of soft attention is central to the advanced models, \textit{Focalis} and \textit{Nexus}, and directly addresses the static bottleneck problem this paper investigates. Further refinements on this grid-based attention include the "Bottom-Up and Top-Down Attention" model by Anderson et al. \cite{anderson2018bottom}, which first identifies salient object regions and then applies top-down attention.

\subsection{The Transformer Revolution and Modern Approaches}
The current state of the art in vision-language tasks has been overwhelmingly shaped by the Transformer architecture \cite{vaswani2017attention}. Models like BLIP \cite{li2022blip} and Flamingo \cite{alayrac2022flamingo} have set new performance benchmarks by employing large-scale, unified Transformer-based frameworks. BLIP, for instance, introduced a multimodal mixture of encoder-decoder models to effectively pre-train on noisy web data, unifying image-text understanding and generation. These models often leverage vision transformers (ViT) \cite{dosovitskiy2020image} for image encoding, dispensing with CNNs entirely.

While attention-based Transformers are dominant, research into non-attention or attention-free architectures continues. For example, RWKV \cite{peng2023rwkv}, a type of Recurrent Neural Network, has been adapted for multimodal tasks, offering a linear-time complexity alternative to the quadratic complexity of standard self-attention. This demonstrates that architectural diversity persists. Unlike the CNN-LSTM models with global average pooling that this paper investigates, architectures like RWKV or vision-specific designs like U-Nets \cite{ronneberger2015u} use different feature processing methods (recurrent states or skip connections) that may mitigate the specific single-vector bottleneck this paper identifies.

\section{An Iterative, Diagnostic Approach to Model Design}
The methodology is rooted in a structured, iterative research process designed to build and refine an image captioning model. This approach provides a crucial, hands-on analysis of the architectural trade-offs, particularly the attention-based resolution of the static information bottleneck that motivated the field's transition towards more complex systems. This paper documents the development of five distinct models, progressively introducing more complex components to address specific limitations and test clear hypotheses.

\subsection{Model 1: Genesis}
The \textit{Genesis} model serves as the foundational baseline, embodying the classic encoder-decoder paradigm that set the stage for modern image captioning. The architecture is intentionally simple to establish a clear performance benchmark. The encoder's role is to "see" the image, for which this paper employs the InceptionV3 architecture pre-trained on ImageNet. This choice provides a strong set of visual features learned from a massive dataset. The final convolutional feature map from InceptionV3 is passed through a Global Average Pooling layer, which condenses the entire spatial grid into a single, fixed-length 2048-dimension feature vector.

This static vector, intended to be a semantic summary of the image, is then fed as the initial hidden state to the decoder. The decoder, a unidirectional Long Short-Term Memory (LSTM) network with 256 units, is responsible for "telling" the story. It generates the caption word-by-word, with each new word conditioned on the previous word and the LSTM's hidden state. The critical limitation, which this model is designed to exemplify, is that the entire burden of representing the image's content rests on this single static vector. This creates a severe \textbf{information bottleneck}, as the decoder has no way to re-examine different parts of the image as it generates the caption. This entire model was trained on the Flickr8k dataset to validate its baseline functionality.

\begin{figure}[h]
    \centering
    \includegraphics[width=0.9\columnwidth]{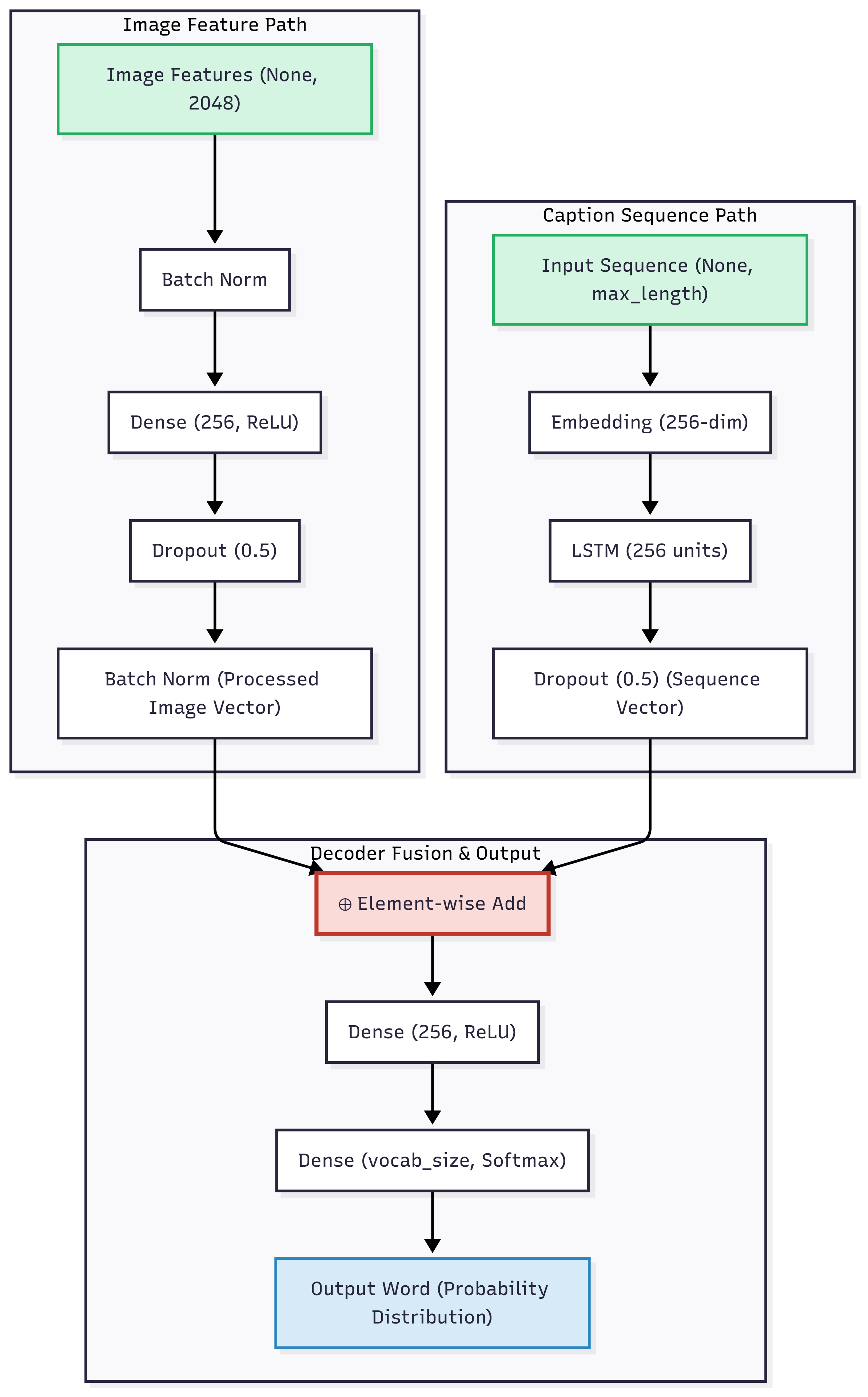}
    \caption{Diagram of the Genesis architecture.}
    \label{fig:Genesis1}
\end{figure}

\subsection{Model 2: Contexta}
While \textit{Genesis} established the visual baseline, its primary limitation was its decoder's simplistic understanding of linguistic structure. A standard unidirectional LSTM only considers past context (the words it has already generated) when predicting the next word. To address this, this paper developed \textit{Contexta}, a model designed to test the hypothesis that a richer language model could improve caption quality, even with the same static visual input.

The core modification was replacing the unidirectional LSTM with a \textbf{Bidirectional LSTM (Bi-LSTM)} with 256 units for each direction. By processing the sequence of previously generated words in both forward and backward directions, the Bi-LSTM creates a more comprehensive understanding of the grammatical and semantic context of the partial caption at each decoding step. This results in a richer 512-dimension language context vector. Furthermore, the fusion mechanism for combining the image vector and the language state was upgraded from simple addition to `Concatenate`. Concatenation preserves the distinct feature representations from both the visual and linguistic streams, allowing subsequent layers to learn more complex interactions between them, whereas simple addition can risk "washing out" information. The goal of \textit{Contexta} was to isolate and measure the impact of a more powerful language decoder.

\subsection{Model 3: Clarity}
The next step was to test a straightforward hypothesis: would a more powerful visual feature extractor directly lead to better captions? To investigate this, this paper designed the \textit{Clarity} model. This paper replaces the InceptionV3 backbone with a state-of-the-art \textbf{EfficientNetV2B3}, known for its superior accuracy. Critically, this paper retained the rest of the \textit{Contexta} architecture, including the global average pooling encoder and the Bi-LSTM decoder.

\begin{figure}[H]
    \centering
    \includegraphics[width=0.8\columnwidth]{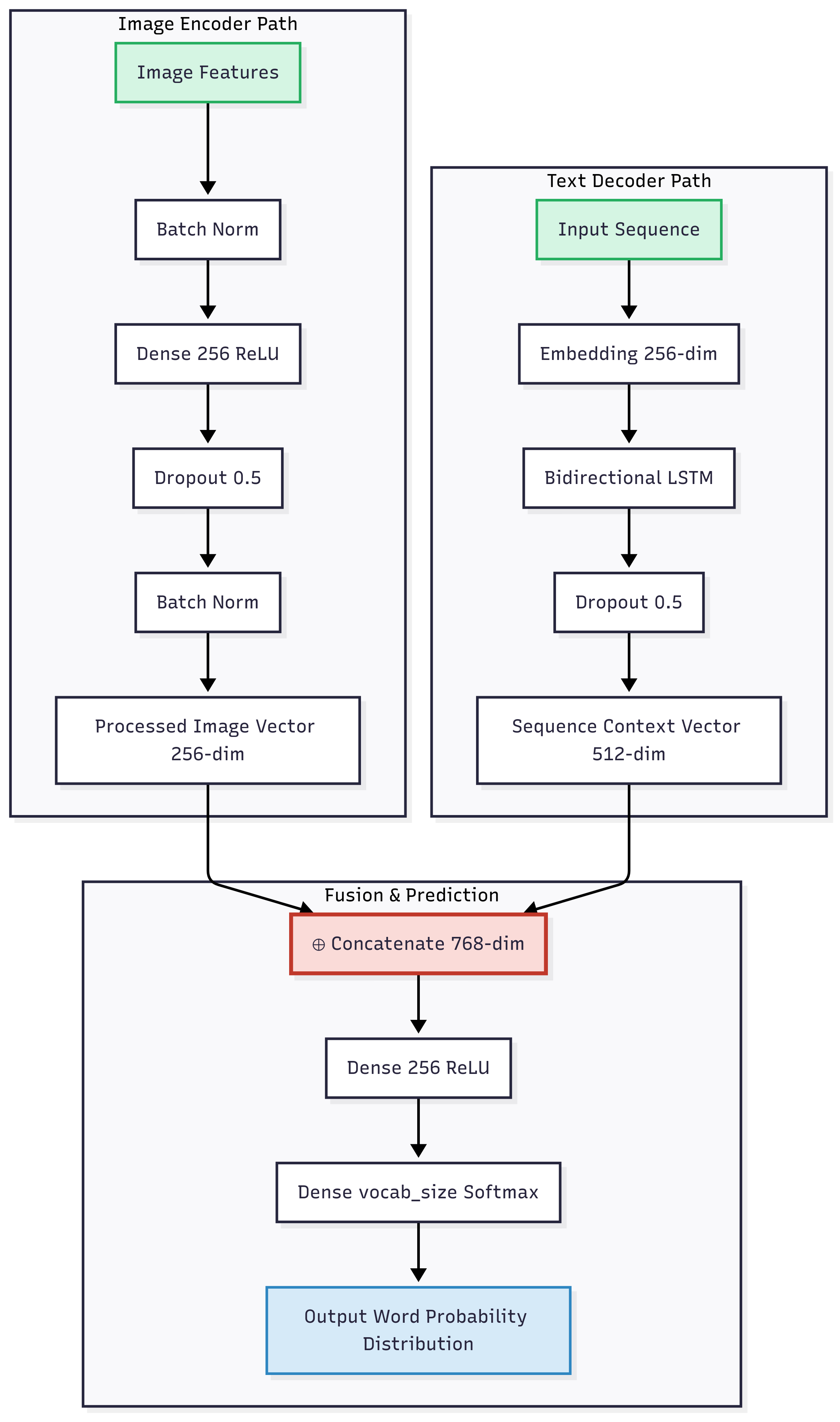}
    \caption{Diagram of the Contexta and Clarity architecture, which uses a static image vector and a Bi-LSTM decoder.}
    \label{fig:Clarity}
\end{figure}

\subsection{Model 4: Focalis}
The performance degradation of \textit{Clarity} provided a critical diagnostic insight: a more powerful visual feature extractor is useless if its output cannot be effectively utilized by the decoder. This confirmed the hypothesis that the single, static feature vector was acting as a severe information bottleneck. Although EfficientNetV2B3 was providing a richer, more detailed map of the image, global average pooling collapses this entire map into a single vector, irretrievably discarding the spatial information that defines "where" objects are. To solve this, this paper designed \textit{Focalis}, which represents a fundamental architectural pivot from a static to a dynamic vision-language interface.

The core innovation in \textit{Focalis} is the introduction of a \textbf{Luong-style attention mechanism}, which completely removes the global average pooling layer. Instead of forcing the model to "remember" the entire image from a single vector, This paper allows it to selectively "look" at different regions of the image at each step of the caption generation process. This enables the decoder to ground specific words in specific spatial locations, for example, focusing on the region containing a "dog" when generating that word, and then shifting focus to the "ball" when describing what the dog is chasing.

The mechanics of the attention process operate at each decoding timestep:
\begin{itemize}
    \item \textbf{Feature Grid:} The EfficientNetV2B3 encoder outputs a (10×10, 1536) feature grid, preserving the spatial layout of the image. After the spatial encoder (described below), this becomes a set of 100 context-aware feature vectors, which serve as the "values" for the attention mechanism.
    \item \textbf{Query, Key, and Value:} At each step, the decoder's current hidden state acts as the "query" vector. This query represents the model's current context, what it has said so far and what it might say next. The query is compared against each of the 100 image patch vectors (the "keys") using an additive scoring function to generate attention weights.
    \item \textbf{Context Vector Generation:} These weights, normalized via a softmax function, determine the importance of each image patch for the current decoding step. A weighted sum of the image patch vectors ("values") is then computed, producing a single, dynamic context vector that is tailored to the specific word being generated. This vector is then concatenated with the decoder's input to predict the next word.
\end{itemize}

Crucially, before the attention mechanism is applied, this paper introduce the key architectural refinement: a \textbf{Bidirectional LSTM as a spatial encoder}. Unlike standard attention mechanisms that treat each image patch independently, this Bi-LSTM processes the 100 patch vectors as a sequence. By reading the sequence of patches forwards and backwards, it enriches each patch's vector with information about its neighbors. For instance, the vector for a patch containing a "face" becomes aware that it is spatially adjacent to a patch containing "hair." This creates a more contextually rich set of feature vectors over which the decoder can attend.

The Bi-LSTM processes the raw feature grid ($h_s$) to produce context-aware vectors ($\bar{h}_s$) that incorporate spatial relationships before they are used by the attention mechanism:
\begin{equation}
    \bar{h}_s = \text{Bi-LSTM}(h_s)
\end{equation}

This complete architecture directly addresses the limitations observed in the previous models. By providing the decoder with dynamic, selective access to context-aware spatial features, this paper hypothesized that \textit{Focalis} would not only resolve the performance degradation seen in \textit{Clarity} but would significantly surpass all prior models. As the results show, this proved to be the case, marking the critical turning point in the development cycle.

\subsection{Model 5: Nexus}
The \textit{Nexus} model does not introduce a new architecture; rather, it represents the culmination of the research by testing the scalability and full potential of the best design. While \textit{Focalis} proved that the attention-based architecture was effective on a smaller scale, \textit{Nexus} was designed to answer a critical question: how does this architecture perform when trained on a large, complex, and diverse dataset?

To this end, this paper took the robust attention-based architecture of \textit{Focalis}, with its EfficientNetV2B3 backbone and Bi-LSTM spatial encoder, and trained it on the large-scale \textbf{MS COCO 2017} dataset. MS COCO, with its ~118k training images, is significantly larger and more complex than Flickr8k. Its images often contain multiple objects in intricate relationships, and it provides five reference captions per image, exposing the model to a much richer range of linguistic expression. The goal of \textit{Nexus} was therefore to unlock the full capabilities of the architecture by leveraging this data. This paper hypothesized that the attention mechanism would learn more nuanced and accurate alignments, the visual encoder would learn more generalizable features, and the language decoder would develop a more sophisticated command of grammar and vocabulary. In essence, \textit{Nexus} serves as the final validation of the iterative design process, demonstrating its effectiveness in a standard benchmark setting.

\begin{figure}[H]
    \centering
    \includegraphics[width=0.9\columnwidth]{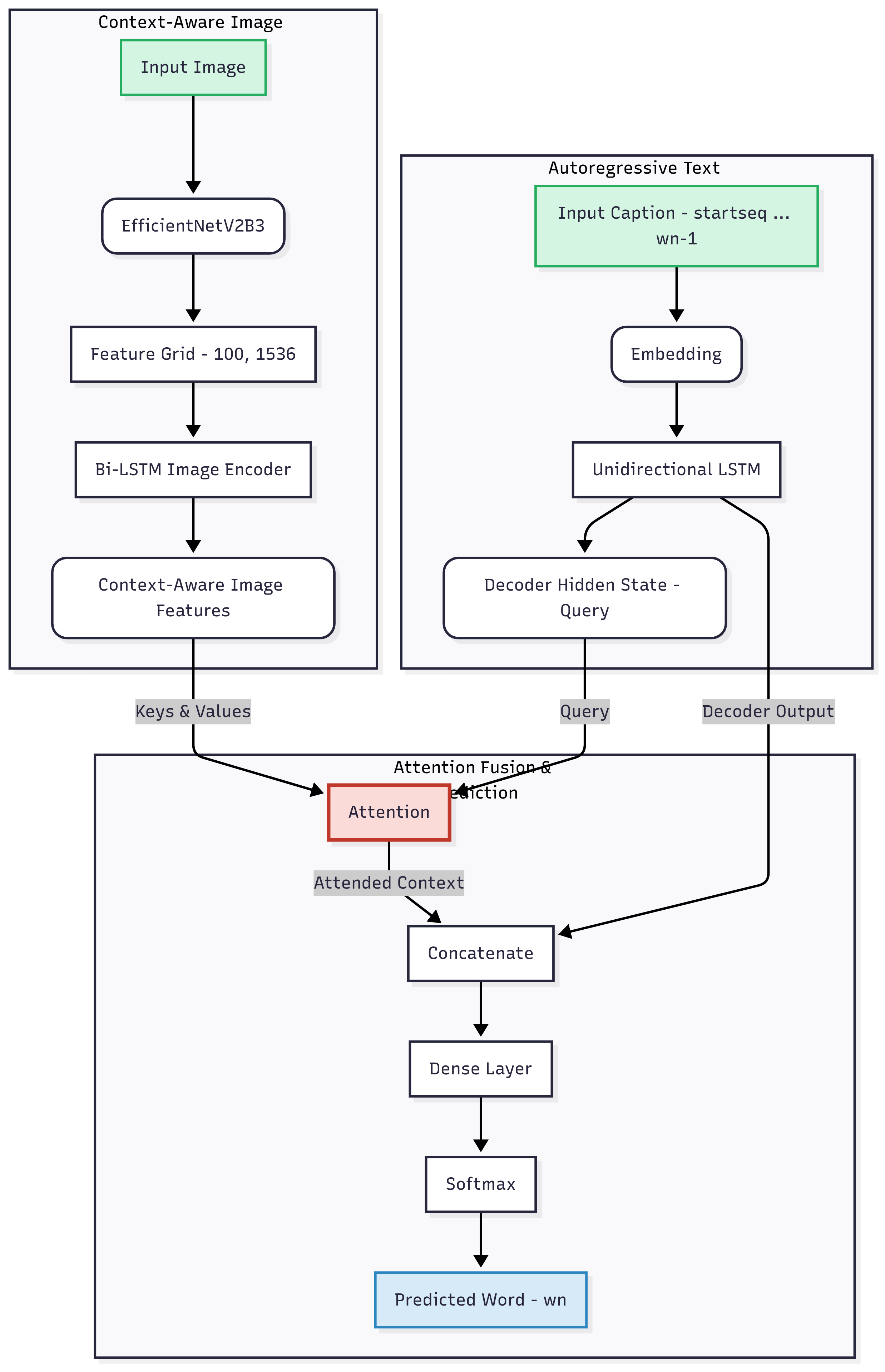}
    \caption{Architecture of the Focalis and Nexus model. This attention-based design allows the decoder to dynamically focus on different image regions for each word.}
    \label{fig:focalis}
\end{figure}

\section{Experiments and Results}
\subsection{Experimental Setup}
\textbf{Datasets:} The initial models (\textit{Genesis}-\textit{Focalis}) were trained on the \textbf{Flickr8k} dataset, which contains approximately 8,000 images. The final model, \textit{Nexus}, was trained on the \textbf{MS COCO 2017} dataset, using the official split of ~118k training images and 5k validation images.

\noindent\textbf{Training:} The \textit{Nexus} model was trained using the Adam optimizer \cite{kingma2014adam} with an initial learning rate of $1 \times 10^{-4}$ and `clipnorm=1.0`. This paper employed a `ReduceLROnPlateau` scheduler (patience=1) and \textit{label smoothing} ($\epsilon=0.1$) with the `CategoricalCrossentropy` loss function. Training was performed using a "teacher forcing" strategy and continued until an `EarlyStopping` callback (patience=3) halted the process after 25 epochs.

\noindent\textbf{Evaluation Metrics:} this paper evaluate model performance using BLEU \cite{papineni2002bleu} and METEOR \cite{banerjee2005meteor}. For inference, this paper use Beam Search with a beam width of $K=7$. As an independent researcher with limited computational resources, the widely-used CIDEr metric was not computed; however, BLEU and METEOR provide a robust evaluation consistent with foundational benchmarks and allow for clear comparison across the iterative models.

\subsection{Result Analysis}
The performance of each model, shown in Table 1, clearly illustrates the impact of the iterative architectural improvements and the critical role of data scaling.

\begin{table}[H]
    \tbl{Performance Progression Across All Models. Models 1-4 are evaluated on Flickr8k; Nexus is evaluated on MS COCO.}{
    \begin{tabularx}{\columnwidth}{@{}llcc@{}}
        \toprule
        \textbf{Model} & \textbf{Dataset} & \textbf{BLEU-4} & \textbf{METEOR} \\
        \midrule
        Genesis & Flickr8k & 14.4 & 26.7 \\
        Contexta   & Flickr8k & 17.2 & 36.1 \\
        Clarity    & Flickr8k & 9.1 & 17.5 \\
        Focalis    & Flickr8k & 20.2 & 38.1 \\
        \midrule
        \textbf{Nexus} & \textbf{MS COCO} & \textbf{31.4} & \textbf{47.4} \\
        \bottomrule
    \end{tabularx}}
    \label{tab:results}
\end{table}

\noindent\textbf{Progressive Enhancement on Flickr8k:} Each architectural step on the smaller dataset provided key insights. The move to a Bi-LSTM in \textit{Contexta} improved scores over \textit{Genesis}. Critically, \textit{Clarity} demonstrated that a more powerful visual backbone (EfficientNetV2) without attention actually degraded performance, exposing the information bottleneck. The introduction of attention in \textit{Focalis} resolved this, achieving the best scores on Flickr8k.

\noindent\textbf{The Power of Scale:} The most dramatic performance leap occurred with the \textit{Nexus} model. By training the best architecture on the large-scale MS COCO dataset, this paper unlocked its true potential. The BLEU-4 score jumped from 20.2 to 31.4, an improvement of over 55\%. This demonstrates that advanced architectures require large, diverse datasets to learn robust, generalizable representations. The final model surpasses several foundational benchmarks, as shown in Table 4.

\noindent\textbf{Qualitative Improvements:} The impact of architectural choices is most strikingly revealed when testing the models on challenging images from the MS COCO dataset, as shown in Table 2. This comparison exposes the combined limitations of both architecture and training data, demonstrating how the information bottleneck in the non-attention models forces them to "hallucinate" when faced with unfamiliar scenes.

The first example is the most telling. When presented with an image of a giraffe, an animal largely absent from the Flickr8k dataset, both \textit{Genesis} and \textit{Clarity} catastrophically misidentify the subject as a "dog". This demonstrates a critical failure: unable to process the unique visual features of the giraffe through their static vector bottleneck, the models are forced to fall back on the statistical priors of their limited training data, where "dog" is a far more common and high-probability subject. They are not describing the image, but rather guessing the most plausible caption from their narrow experience.

This pattern of failure extends to other complex scenes. For the image of kites flying, the non-attention models again hallucinate an entirely different, high-probability scenario involving "people" at a "beach" or in "water," defaulting to generic scenes prevalent in their training set rather than parsing the specific objects present. Similarly, for the dining image, they fail to grasp the specific context, demonstrating a failure of both object recognition and semantic relationship understanding when the scene deviates from their training experience.

In stark contrast, \textit{Nexus}, whose architecture was designed to overcome this bottleneck and was trained on the richer and more diverse MS COCO dataset, correctly identifies the "giraffe," the "kites," and the specific dining scene. This powerful comparison highlights that a superior architecture (the attention mechanism) \textit{combined with} large-scale, relevant data is fundamental to achieving accurate semantic grounding and moving beyond simple pattern matching. 

\begin{table}[H]
    \tbl{Qualitative Comparison of Sample Captions.}{
    \begin{tabularx}{\columnwidth}{@{} l p{0.80\columnwidth} @{}}
        \toprule
        \textbf{Model} & \textbf{Generated Caption} \\ 
        \midrule
        \multicolumn{2}{@{}l}{\textit{Image: A giraffe standing in a field (see Appendix Fig. 6)}} \\
        Genesis & A brown and white dog is running through the grass. \\
        Clarity & A dog is running through a grassy field. \\
        Nexus   & A giraffe standing on top of a lush green field. \\
        \midrule
        \multicolumn{2}{@{}l}{\textit{Image: Kites flying in the sky (see Appendix Fig. 7)}} \\
        Genesis & A group of people are playing in the water. \\
        Clarity & A group of people are walking on the beach. \\
        Nexus   & There are many kites flying in the sky. \\
        \midrule
        \multicolumn{2}{@{}l}{\textit{Image: People sitting at a table (see Appendix Fig. 8)}} \\
        Genesis & A group of people are standing in front of a crowd of people. \\
        Clarity & A group of people are sitting at a restaurant. \\
        Nexus   & Several people sitting around a table with plates of food. \\
        \bottomrule
    \end{tabularx}}
    \label{tab:qualitative}
\end{table}

\noindent\textbf{Training Dynamics and Model Selection}
The training process for \textit{Nexus} revealed a divergence between the validation loss (the optimization target) and the final evaluation metrics. As shown in Figure 4, \texttt{val\_loss} continued to improve until Epoch 25. However, an evaluation of checkpoints on a random 15-image sample (Table 3) showed that peak performance on metrics like F1-Score and BLEU-4 occurred earlier, around Epoch 13, where \texttt{val\_loss} and \texttt{train\_loss} first converged. After this point, the model began to over-optimize on the cross-entropy loss at the expense of caption quality. This finding is consistent with observations from prior work \cite{xu2015show}, where a breakdown in correlation between log-likelihood and BLEU was also noted. This analysis led us to select the checkpoint from \textbf{Epoch 13 as the champion model} for final evaluation and reporting.

\begin{table}[H]
    \tbl{Metric Comparison of Nexus Checkpoints on a Sample Set.}{
    \resizebox{\columnwidth}{!}{%
    \begin{tabular}{@{}lccc@{}}
        \toprule
        \textbf{Metric} & \textbf{Epoch 10} & \textbf{Epoch 13} & \textbf{Epoch 25} \\
        \midrule
        BLEU-1     & \textbf{0.8352} & 0.7979 & 0.6066 \\
        BLEU-2     & 0.6961 & \textbf{0.6962} & 0.4643 \\
        BLEU-3     & 0.5428 & \textbf{0.5661} & 0.3294 \\
        BLEU-4     & 0.4192 & \textbf{0.4650} & 0.1856 \\
        METEOR     & 0.5557 & 0.5869 & \textbf{0.5878} \\
        Precision  & \textbf{0.8571} & 0.8367 & 0.8495 \\
        Recall     & 0.7372 & \textbf{0.7750} & 0.6049 \\
        F1-Score   & 0.7927 & \textbf{0.8047} & 0.6639 \\
        \bottomrule
    \end{tabular}%
    }}
    \label{tab:epoch_comparison}
\end{table}

The analysis of the training dynamics highlights a significant divergence between the optimization objective (validation loss) and the task's primary evaluation metrics (e.g., BLEU, METEOR). This difference arises because the CategoricalCrossentropy loss function serves as an imperfect proxy for sentence-level quality. While it effectively guides the model in next-word prediction, it does not explicitly reward n-gram overlap or semantic coherence across an entire generated sequence. This paper observe that after reaching a point of convergence around Epoch 13, further training leads to overfitting on this proxy objective. This results in the model favoring high-probability, often repetitive, phrases that continue to reduce loss but degrade overall caption quality, as reflected by decreasing BLEU scores. This misalignment between log-likelihood maximization and human-centric evaluation metrics is a recognized challenge in sequence generation, as previously discussed by Xu et al. \cite{xu2015show}.

Consequently, although the model at Epoch 25 achieves the minimum validation loss, the checkpoint from Epoch 13 represents the optimal model for the task. It achieves the highest scores on metrics like BLEU-4 and METEOR, which are more closely aligned with human judgments of quality. This finding demonstrates the importance of using a comprehensive suite of task-specific metrics for model selection, rather than relying solely on the training loss.
\clearpage
\begin{figure*}[t!]
    \centering
    \includegraphics[width=\textwidth]{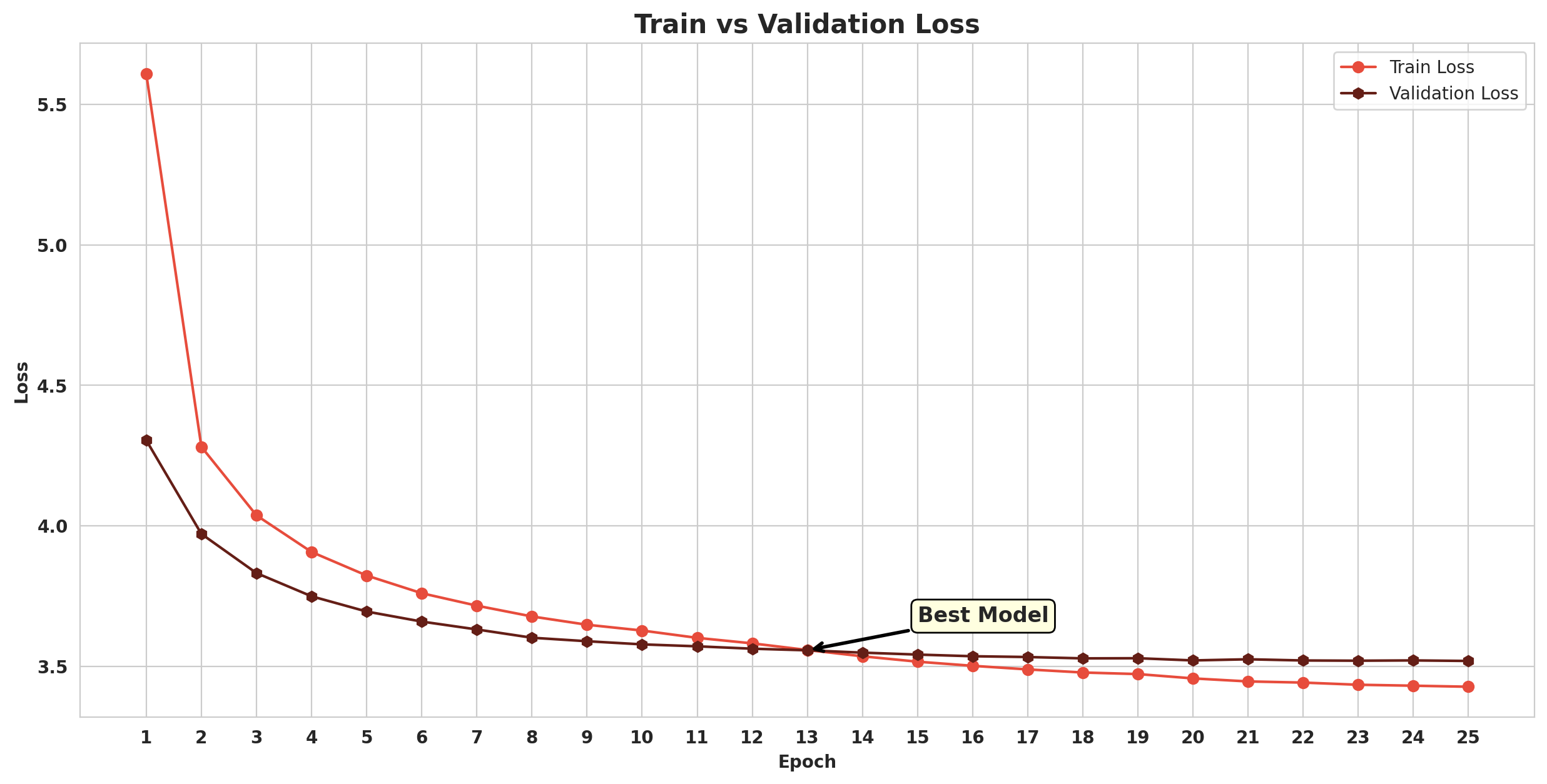} 
    \caption{Training and validation loss curves for the Nexus model over 25 epochs.}
    \label{fig:loss}
\end{figure*}

\vspace{2mm}

\begin{figure*}[t!]
    \centering
    \includegraphics[width=\textwidth]{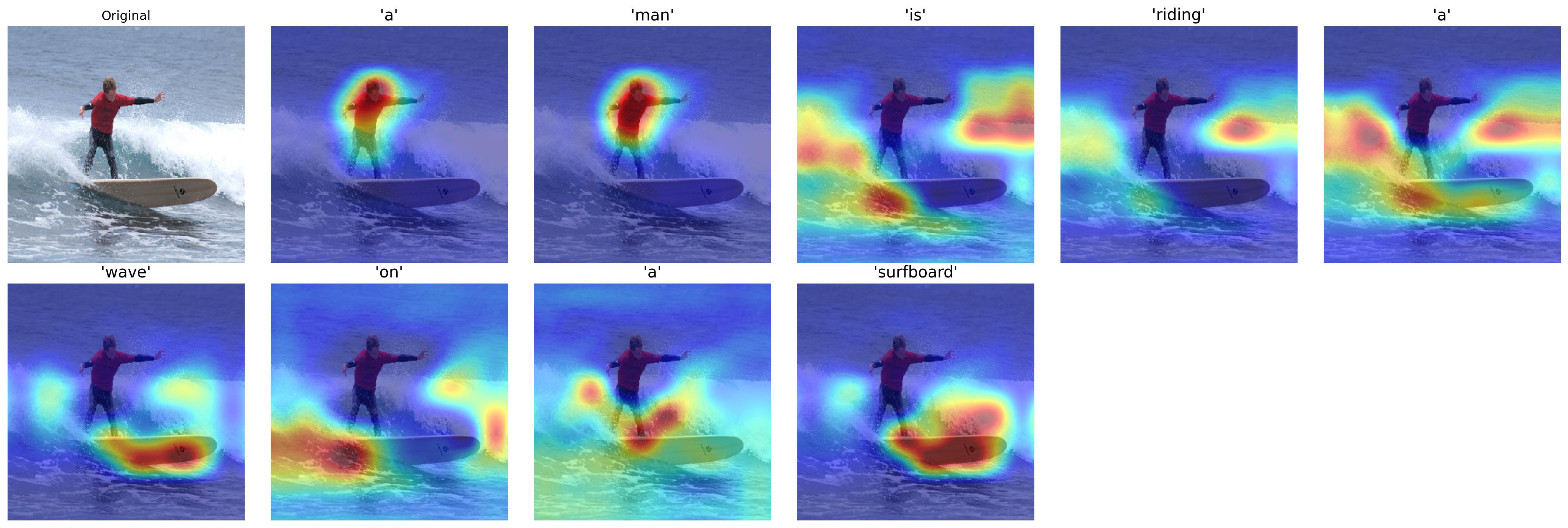}
    \begin{minipage}{\textwidth}
        \vspace{0.5em}
        \centering
        \small
        \textbf{Epoch 10:} A person riding a surfboard on a wave. \\
        \textbf{Epoch 13:} A man is riding a wave on a surfboard. \\
        \textbf{Epoch 25:} A man riding a wave on top of a surfboard.
    \end{minipage}
    \caption{Attention visualization for a sample image using the Nexus model checkpoint from \textbf{Epoch 13}. The heatmaps show the model’s focus shifting across image regions as it generates each word in the caption. Captions from Epochs 10 and 25 are included below the image for comparison, but only the Epoch 13 caption corresponds to the attention map shown.}
    \label{fig:attention_map}
\end{figure*}
\clearpage

\section{Conclusion}
In this paper, this paper have detailed the iterative design and evaluation of five image captioning models, charting a course from a simple baseline to a competitive attention-based system. The systematic approach demonstrates a clear path for model enhancement and yields several key insights. First, this paper show that within the classic CNN-LSTM paradigm, simply upgrading a visual backbone is not a guaranteed path to better performance; it must be paired with an architecture, such as an attention mechanism, that can effectively process the increased feature complexity. Second, this paper show that the point of lowest validation loss does not always correspond to peak performance on task-specific metrics like BLEU, necessitating a nuanced model selection process.

While the final model, \textit{Nexus}, does not compete with the scale or performance of contemporary Transformer-based systems like BLIP, its final BLEU-4 score of 31.4 validates the architectural design by surpassing several foundational benchmarks. This work's contribution is not a new state-of-the-art result, but rather a clear, educational, and practical guide to understanding the foundational trade-offs in vision-language architecture that precipitated the field's shift toward modern, attention-centric designs.

\subsection*{Future Work}
While the \textit{Nexus} model demonstrates strong performance, several areas remain open for exploration. As an independent researcher, this work was developed with limited computational resources, which motivated the focus on efficient CNN-LSTM architectures and guided the evaluation choices. Future work could investigate lightweight or quantized variants of the architecture for deployment in resource-constrained environments. Second, generalization to out-of-distribution or low-resource image domains warrants deeper investigation, potentially using domain adaptation or contrastive pre-training techniques.

\begin{table*}[ht]
\centering
\caption{Final Benchmark Comparison on the MS COCO Validation Set. Scores are reported as percentages. The Nexus model is compared against foundational models and modern Transformer-based systems to provide a clear context for its performance.}
\label{tab:sota_comparison}
\begin{tabularx}{\textwidth}{lccccc}
    \toprule
    \textbf{Model} & \textbf{BLEU-1} & \textbf{BLEU-2} & \textbf{BLEU-3} & \textbf{BLEU-4} & \textbf{METEOR} \\
    \midrule
    \multicolumn{6}{l}{\textit{Foundational CNN-LSTM Models}} \\
    BRNN (Karpathy \& Li, 2015) \cite{karpathy2015deep} & 64.2 & 45.1 & 30.4 & 20.3 & -- \\
    Google NIC (Vinyals et al., 2015) \cite{vinyals2015show} & 66.6 & 46.1 & 32.9 & 24.6 & -- \\
    Soft-Attention (Xu et al., 2015) \cite{xu2015show} & 70.7 & 49.2 & 34.4 & 24.3 & 23.9* \\
    Hard-Attention (Xu et al., 2015) \cite{xu2015show} & 71.8 & 50.4 & 35.7 & 25.0 & 23.0* \\
    Up-Down (Anderson et al., 2018) \cite{anderson2018bottom} & 79.8 & -- & -- & 36.3 & 27.7* \\
    \midrule
    \textbf{Nexus (This work, Epoch 13)} & \textbf{71.7} & \textbf{55.0} & \textbf{41.8} & \textbf{31.4} & \textbf{47.4} \\
    \midrule
    \multicolumn{6}{l}{\textit{Transformer-Based Models (for reference)}} \\
    OSCAR (Li et al., 2020) \cite{li2020oscar} & -- & -- & -- & 36.5 & 33.5* \\
    BLIP (Li et al., 2022) \cite{li2022blip} & -- & -- & -- & 39.7 & 36.1* \\
    \bottomrule
\end{tabularx}

\vspace{2mm}
\begin{minipage}{\textwidth}
\footnotesize
\textit{*METEOR scores from different eras/packages may not be directly comparable. The NLTK-based METEOR score is reported for completeness.}
\end{minipage}
\end{table*}

\appendix
\section{Qualitative Examples and Visualizations}
\label{sec:appendix}

This appendix provides additional visual results to complement the quantitative analysis in the main paper. This paper showcase attention maps from the best model, \textit{Nexus}, to illustrate how the model dynamically focuses on relevant image regions during caption generation.

\begin{figure}[H] 
    \centering
    \includegraphics[width=0.8\columnwidth]{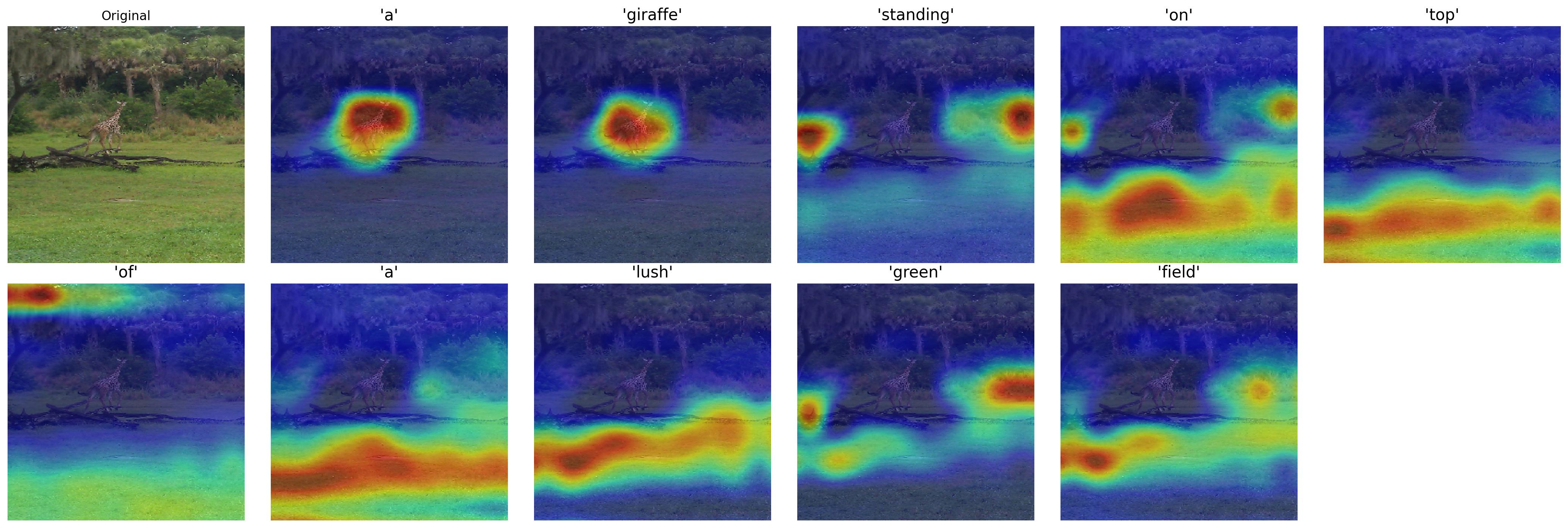}
    \caption{A giraffe standing on top of a lush green field.}
    \label{fig:appendix_attn_map_1}
\end{figure}
\begin{figure}[H] 
    \centering
    \includegraphics[width=0.8\columnwidth]{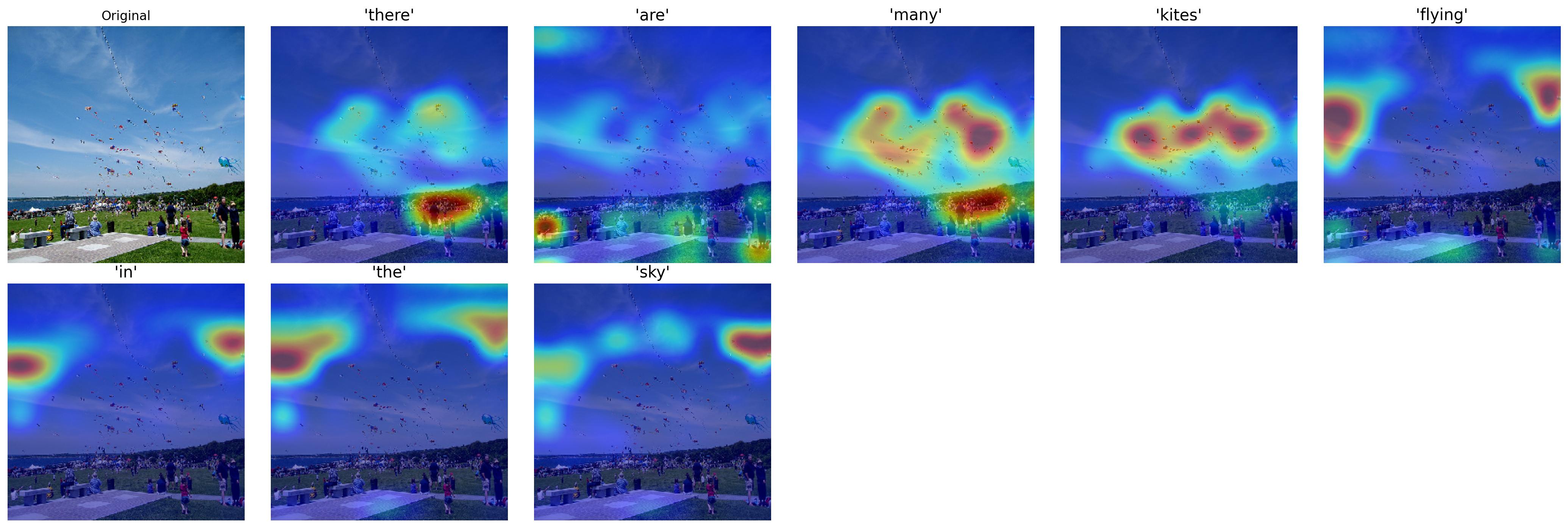} 
    \caption{There are many kites flying in the sky.}
    \label{fig:appendix_attn_map_2}
\end{figure}
\begin{figure}[H] 
    \centering
    \includegraphics[width=0.8\columnwidth]{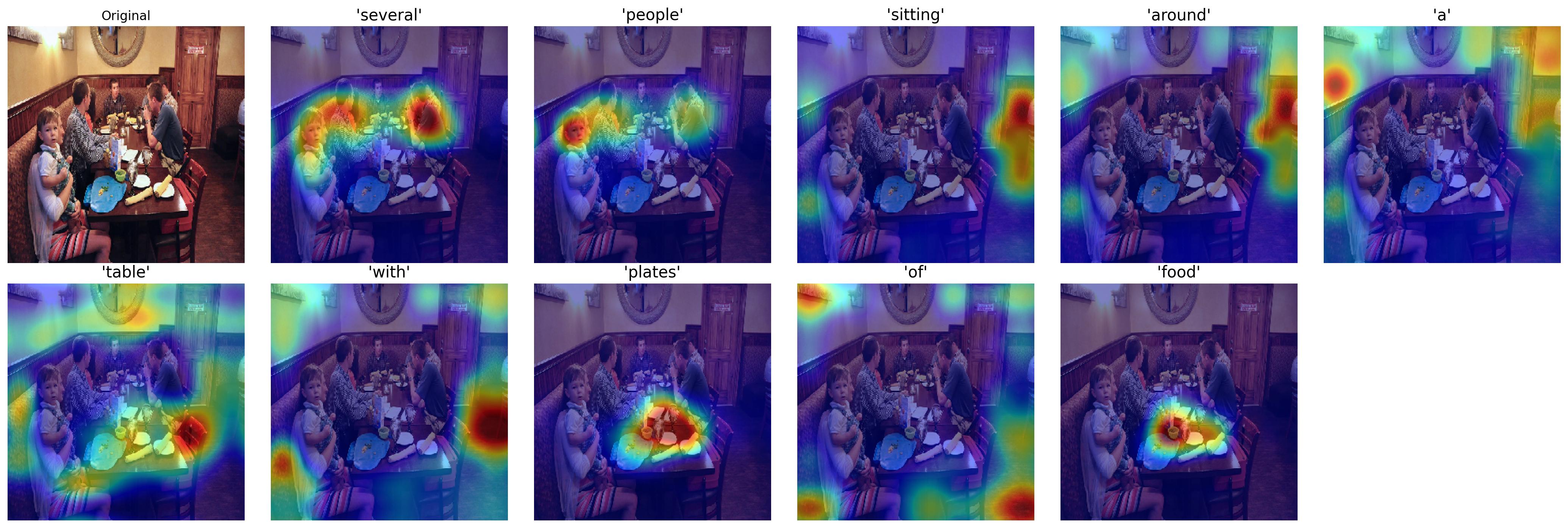}
    \caption{Several people sitting around a table with plates of food.}
    \label{fig:appendix_attn_map_3}
\end{figure}
\begin{figure}[H] 
    \centering
    \includegraphics[width=0.8\columnwidth]{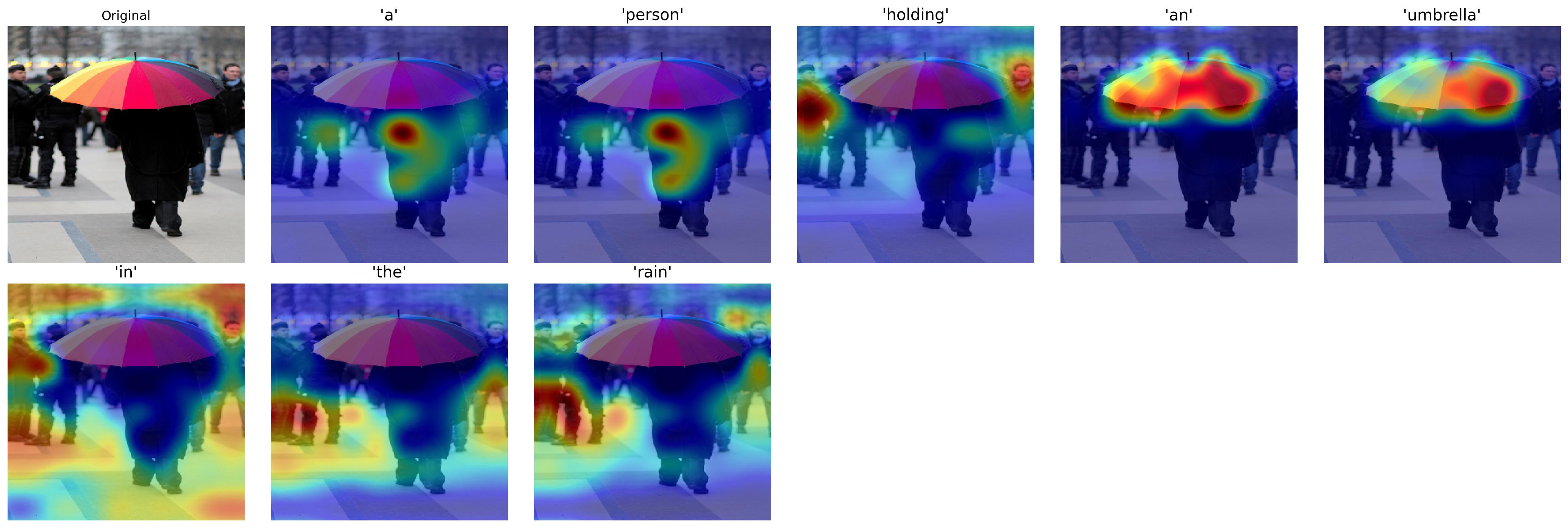} 
    \caption{A person holding an umbrella in the rain.}
    \label{fig:appendix_attn_map_5}
\end{figure}
\begin{figure}[H]
    \centering
    \includegraphics[width=0.8\columnwidth]{000000392933.jpg} 
    \caption{\textbf{Nexus Caption:} "A giraffe standing on top of a lush green field."}
    \label{fig:appendix_giraffe}
\end{figure}
\begin{figure}[H]
    \centering
    \includegraphics[width=0.8\columnwidth]{000000361730.jpg}
    \caption{\textbf{Nexus Caption:} "There are many kites flying in the sky."}
    \label{fig:appendix_kites}
\end{figure}
\begin{figure}[H] 
    \centering
    \includegraphics[width=0.8\columnwidth]{000000404678.jpg}
    \caption{\textbf{Nexus Caption:} "Several people sitting around a table with plates of food."}
    \label{fig:appendix_dining}
\end{figure}

\end{document}